\documentclass[conference]{IEEEtran}

\usepackage[english]{babel}
\usepackage[T1]{fontenc}
\usepackage[utf8]{inputenc}   
\usepackage{csquotes}

\usepackage{amsmath,amssymb,amsthm}
\usepackage{mathtools}

\usepackage{graphicx}
\usepackage{booktabs}               
\usepackage{makecell}
\usepackage{array}
\usepackage{float}                 
\usepackage{placeins}               
\usepackage{dblfloatfix}
\usepackage[caption=false,font=normalsize,labelfont=sf,textfont=sf]{subfig}

\usepackage[hidelinks]{hyperref}   
\usepackage[nameinlink,capitalize,noabbrev]{cleveref} 

\theoremstyle{plain}

\theoremstyle{definition}

\theoremstyle{remark}

\title{Liquid Neural Network Models for Natural Gas Spot Price Time-Series Forecasting}

\author{
  \IEEEauthorblockN{Yiqian Liu, Jiayi Niu, Adam Kelleher, Subhabrata Das$^{*}$\thanks{$^{*}$Corresponding author: Subhabrata Das (email: sd2957@columbia.edu)}}
  \IEEEauthorblockA{
    \textit{Department of Data Science, Columbia University} \\
    New York, NY, USA \\
    *Corresponding author: sd2957@columbia.edu\\
  }
}

\begin{document}

\maketitle
\begin{abstract}
Natural gas is undoubtedly an essential component of the global energy system. Accurate short-term forecasting of natural gas price is challenging due to pronounced volatility driven by seasonal demand patterns, geopolitical developments, and shifting macroeconomic conditions. The nonlinear dynamics and frequent regime changes can limit the effectiveness of traditional time-series models. In this study, we explore the use of Liquid Neural Networks (LNNs) for short-horizon forecasting of the Henry Hub spot price, a primary benchmark for pricing. LNNs are designed to adapt continuously to evolving temporal patterns through dynamic internal state updates, making them well suited for nonstationary price behavior. By improving forecast accuracy in volatile market conditions, this work aims to reduce uncertainty and enhance decision support across energy trading and power market applications.
\end{abstract}

\vspace{0.5em}

\section{Introduction}
\subsection{Background and Motivation}
Natural gas remains a cornerstone of the global energy matrix, serving as a
critical bridge fuel in the transition toward low-carbon energy systems while
providing baseload reliability for power generation \cite{safari2019}. In the
United States, the Henry Hub spot price functions as the primary benchmark
for futures, options, and physical trading \cite{mazighi2005}, making its
predictability of paramount importance to utility operators, energy traders,
and financial institutions alike.

The natural gas market is characterized by pronounced volatility and
nonstationary behavior driven by weather shocks, geopolitical developments,
and supply disruptions \cite{hailemariam2019}. Traditional econometric models
such as ARIMA and GARCH offer statistical interpretability, but are limited by
linear assumptions and stationarity conditions during periods of structural
change \cite{aslan2011}. Rolling-window OLS addresses some of this rigidity
through adaptive re-estimation, yet remains constrained by its linear
functional form and cannot capture the nonlinear dependencies and
regime-dependent volatility clustering characteristic of natural gas markets.
Deep learning approaches, particularly LSTM networks, improve on this by
modeling nonlinear temporal dependencies, but their fixed internal dynamics
limit responsiveness when market regimes shift rapidly \cite{ilhan2020}.
Thus, many existing approaches rely on fixed or discretely updated parameters,
limiting their ability to continuously adapt to evolving market conditions. This study examines whether adaptivity improves short-term forecast accuracy compared to rolling-window linear regression. 
\subsection{Research Scope and Contributions}
To address this gap, we formulate Henry Hub price forecasting as a
nonstationary time-series prediction problem and investigate Liquid Neural
Networks (LNNs) — recurrent architectures whose dynamic time constants allow
internal states to adapt continuously to incoming information — as a
candidate solution. We construct a ten-and-a-half-year daily dataset
(January 2015–August 2025) integrating Henry Hub spot prices with financial,
energy, and operational variables, and focus on next-day short-horizon
forecasting rather than long-term price projection.

This study makes three principal contributions. First, we conduct a
systematic comparison of five liquid neural architectures — LTC, Strict CfC,
Hybrid CfC, CT-LTC, and a standard LSTM benchmark — against a
rolling-window linear regression baseline for short-horizon Henry Hub
forecasting. Second, we employ a stratified expanding-window evaluation
protocol with Moving Block Bootstrap uncertainty quantification to produce
statistically reliable performance estimates that account for temporal
autocorrelation in nonstationary forecast errors. Third, through controlled
architectural comparison, we identify which design characteristics —
specifically, input-conditioned timescale modulation versus fixed or
calendar-based step sizing — drive performance differences under the
high-frequency regime shifts that typify natural gas markets.

\section{Related Work}
Forecasting energy prices has been widely studied using both statistical and machine learning approaches. Early econometric models, such as autoregressive and cointegration frameworks, effectively captured short-term correlations but struggled during periods of structural market change caused by supply disruptions or policy shifts. Later, neural models such as LSTM and GRU improved nonlinear pattern recognition but often at the expense of interpretability and computational stability.
Several studies have proposed rolling-window regression frameworks to better handle time-varying relationships. Dbouk and Jamali introduced a rolling estimation approach for daily crude oil prices, where model parameters are re-estimated within a fixed-length moving window to adapt to recent market conditions while discarding outdated information \cite{dbouk2018}. Artemova \textit{et al.} and Khan \textit{et al.} similarly demonstrate that rolling-window linear regression provides a strong baseline for short-term forecasting of volatile energy commodities \cite{artemova2025,khan2019}, supporting the principle that rolling estimation enhances robustness and responsiveness — a design choice we adopt in our baseline model.
The challenge of applying deep learning to volatile energy time-series has motivated a range of hybrid and adaptive architectures. Das \textit{et al.} propose a PCA-PR-Seq2Seq-Adam-LSTM framework for power outage prediction, demonstrating that structured preprocessing and temporal architectures substantially improve robustness under abrupt distributional shifts analogous to Henry Hub regime changes \cite{das2025outage}. Multi-headed recurrent networks tuned via meta-heuristic optimization have also been applied to energy price forecasting, reinforcing the broader insight that models adapting to evolving dynamics outperform those with fixed parameters \cite{springer2025}.
The study most closely related to ours is Liu \textit{et al.} \cite{liu2025plos}, who compare feedforward networks, support vector machines, random forests, and LSTM for multi-step Henry Hub forecasting. Their results establish LSTM as a strong neural baseline and highlight systematic accuracy degradation with increasing horizon. However, their recurrent dynamics remain governed by fixed time scales, which may limit responsiveness under abrupt regime shifts. Zaragoza \textit{et al.} address adaptivity from a different angle, proposing a time-varying graph learning framework for electricity price forecasting in which inter-node dependencies evolve dynamically \cite{zaragoza2024}, underscoring a broader trend toward adaptive modeling in energy markets.
Building on this literature, we introduce Liquid Neural Networks — recurrent architectures designed to capture fast-changing temporal dependencies through adaptive internal dynamics with fewer parameters and improved stability compared to standard recurrent models \cite{hasani2021ltc,hasani2022}, making them particularly suitable for the volatile, regime-prone behavior of Henry Hub spot prices.
\section{Dataset}
\subsection{Data}
We utilize Henry Hub Natural Gas Spot Prices \cite{eia_ng} as the target variable. Since gas pricing is driven by a mix of supply, demand, and market conditions, we construct a predictor set combining fundamentals and market indicators. WTI crude oil spot prices \cite{eia_wti} serve as a benchmark for broader energy market trends and cross-commodity substitution. U.S. Treasury yield curve rates \cite{treasury} — specifically the 2-year, 5-year, 10-year, and 20-year maturities — capture macroeconomic and capital cost conditions relevant to natural gas investment and trading. The U.S. Dollar Index \cite{investing_usd} reflects dollar strength, which influences the competitiveness of U.S. LNG exports. Equity-based measures, including the S\&P Energy Index \cite{investing_snp} and the Dow Jones U.S. Coal Index \cite{investing_coal}, provide signals of sector-wide sentiment and coal--gas substitution effects. Firm-level production sentiment is proxied using daily stock prices of EQT \cite{investing_eqt}. Finally, U.S. nuclear generation data — represented by installed capacity, outage capacity (both in megawatts), and percentage outage — reflects shifts in electricity supply capacity where reductions in nuclear availability can increase residual demand for natural gas \cite{eia_nuclear}.
All datasets were merged by date over the period January~6,~2015 to August~29,~2025. Variables with a high proportion of missing observations — specifically Coal Volatility, USD Index Volatility, the CBOE Crude Oil ETF Volatility Index (VIX), Treasury 1.5 Month, Treasury 4 Month, S\&P Energy Volatility, Treasury 2 Month, and Treasury 30-Year rates — were excluded due to insufficient daily coverage. At the observation level, 54 rows were removed from the original 2,699 daily observations (approximately 2.0\%), primarily reflecting missing values associated with market holidays, and including three observations where the spot price level exceeded 5.5 standard deviations from the median. Large single-day return spikes are deliberately retained, as they represent genuine market events — most notably the February 2021 Winter Storm Uri disruption — and removing them would introduce survivorship bias into the forecasting evaluation.

The final dataset spans January 6, 2015 to August 29, 2025, contains 2,645 observations and 30 predictor features for the standard neural models (29 for the CT-LTC variant, which additionally excludes the lagged return to avoid target leakage), with the daily natural gas spot price (USD per MMBtu) as the target variable.

\subsection{Exploratory Data Analysis}

\FloatBarrier

To characterize the empirical structure of the dataset, we conduct exploratory analyses of the Henry Hub natural gas spot price and selected energy and financial market indicators over the period 2015--2025.
\subsubsection{Henry Hub Natural Gas Spot Price Dynamics}
\begin{figure}[H]
    \centering
    \begin{minipage}[t]{0.48\textwidth}
        \centering
        \includegraphics[height=3.95cm]{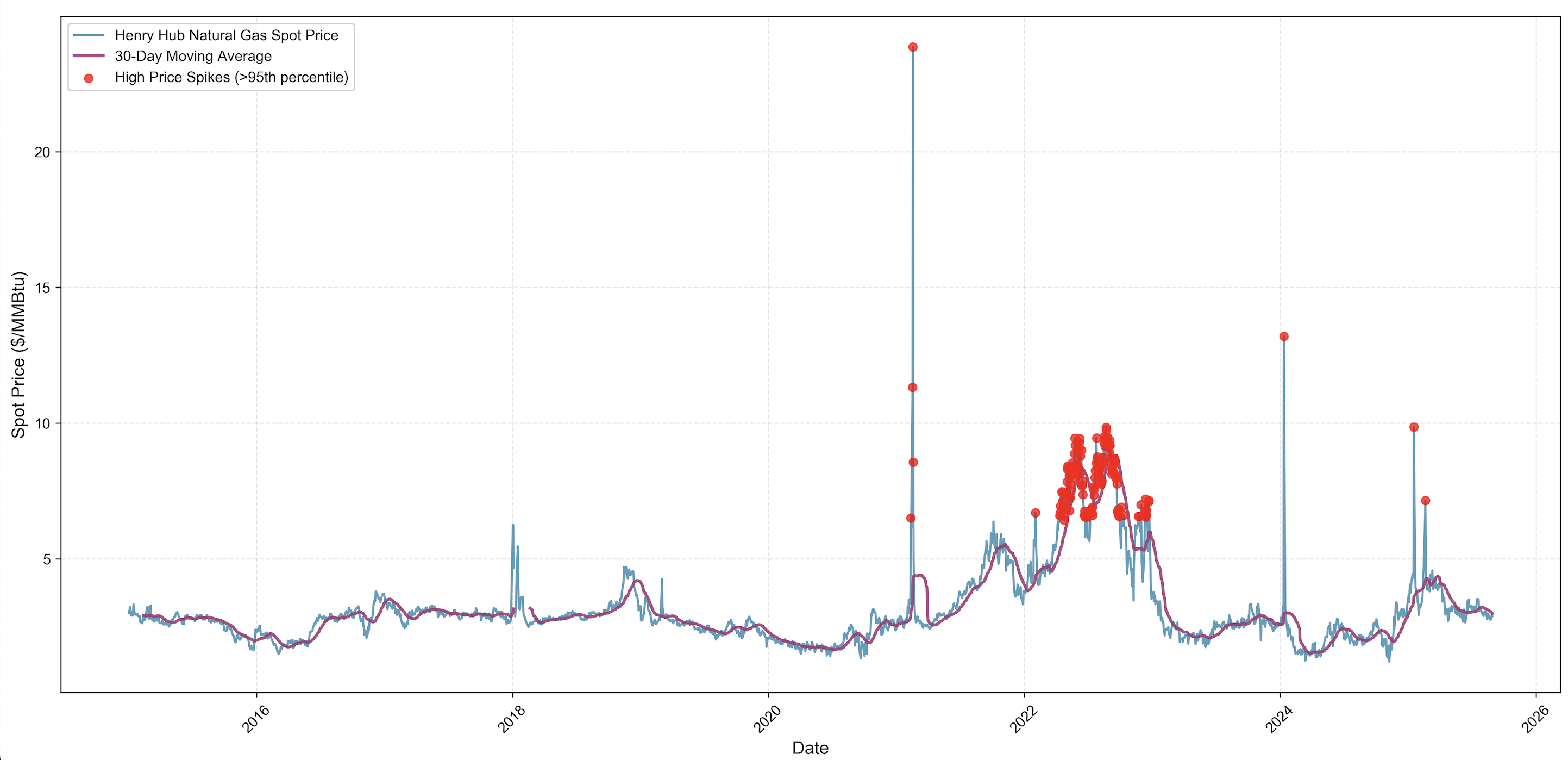}
        \caption{Henry Hub Natural Gas Spot Price Trend (2015--2025)}
        \label{fig:henry_hub_trend}
    \end{minipage}
\end{figure}

Figure~\ref{fig:henry_hub_trend} presents the daily Henry Hub natural gas spot price together with a 30-day moving average, with observations exceeding the 95th percentile highlighted to emphasize periods of extreme price behavior. The series exhibits pronounced volatility and distinct regime changes, with particularly large price dislocations during 2021--2022, reflecting heightened stress in global energy markets.

\subsubsection{Energy and Financial Market Performance}
\begin{figure}[H]
    \begin{minipage}[t]{0.45\textwidth}
        \centering
        \includegraphics[height=3.95cm]{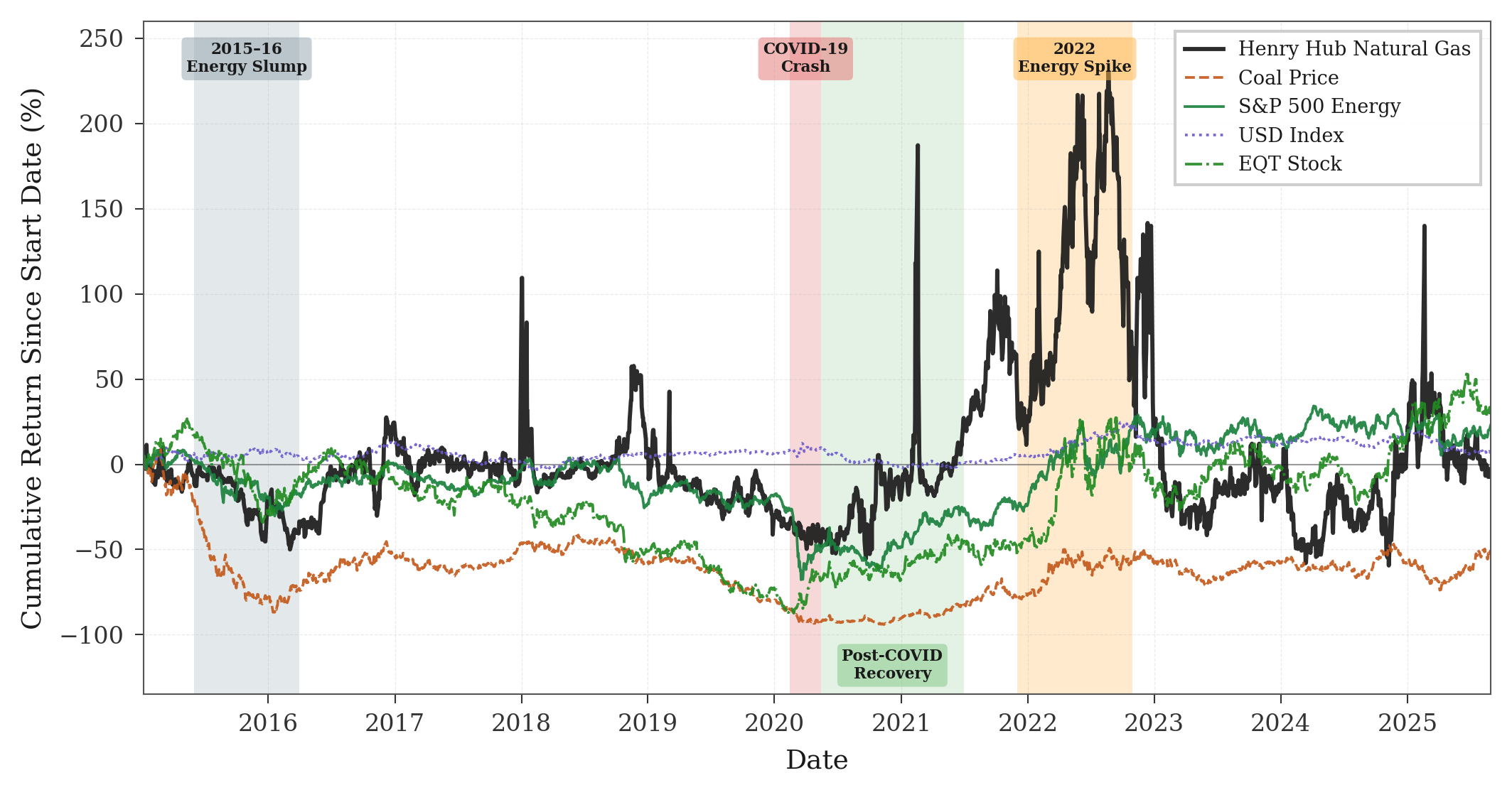}
        \caption{Energy and Financial Market Performance (2015--2025)}
        \label{fig:energy_finance_performance}
    \end{minipage}
\end{figure}

Figure~\ref{fig:energy_finance_performance} places Henry Hub in the energy and finance context. Compared to other series, natural gas cumulative returns (black line) exhibit extreme volatility, ranging from -80\% to +220\%, while the USD Index (purple) barely moves beyond ±15\%. Several structural observations stand out. Coal (orange dashed) has lost 60–80\% of its 2015 starting value. S\&P 500 Energy (green) and EQT (green dash-dot) track each other well, collapsing during the COVID-19 pandemic and recovering in 2021–2022, supporting the heatmap's strong inter-feature correlations. Energy commodity moves are least correlated with the USD Index, the most stable series. The shaded event windows show how external shocks change the return landscape: the COVID crash erased cumulative gains across all series, the post-COVID recovery saw divergent recoveries (energy equities rebounded while coal did not), and the 2022 energy spike produced the largest natural gas returns in the sample due to the Russia-Ukraine conflict and European supply disruptions.

\subsubsection{Feature Correlations}
\begin{figure}[H]
    \centering
    \includegraphics[width=0.47\textwidth]{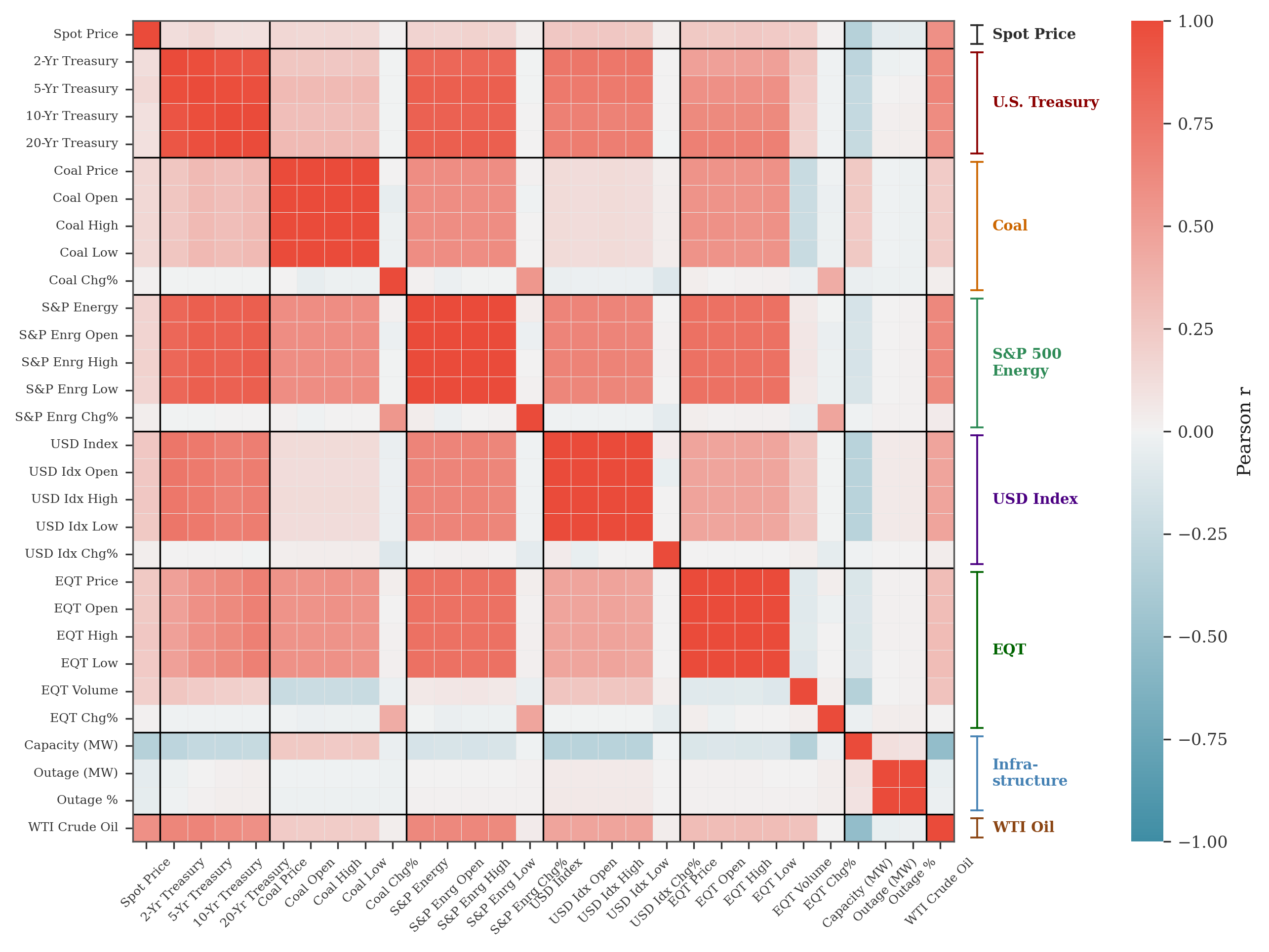}
    \caption{Correlation Heatmap}
    \label{fig:feature_correlation}
\end{figure}

Figure~\ref{fig:feature_correlation} shows the pairwise Pearson correlation structure between the Henry Hub spot price and selected predictor variables. The heatmap shows weak linear correlations between the spot price and most individual predictors, with the exception of capacity (MW) and WTI crude oil spot prices. In contrast, several predictors show strong inter-feature correlations, particularly U.S. Treasury yields, coal prices, energy equity indices, the US Dollar Index, and EQT stock prices. These patterns demonstrate the presence of multicollinearity and complex cross-market relationships, providing empirical context for the forecasting problem addressed in this study.

\subsubsection{Rolling 30-Day Volatility}
\begin{figure}[H]
    \centering
    \begin{minipage}[t]{0.48\textwidth}
        \centering
        \includegraphics[height=3cm]{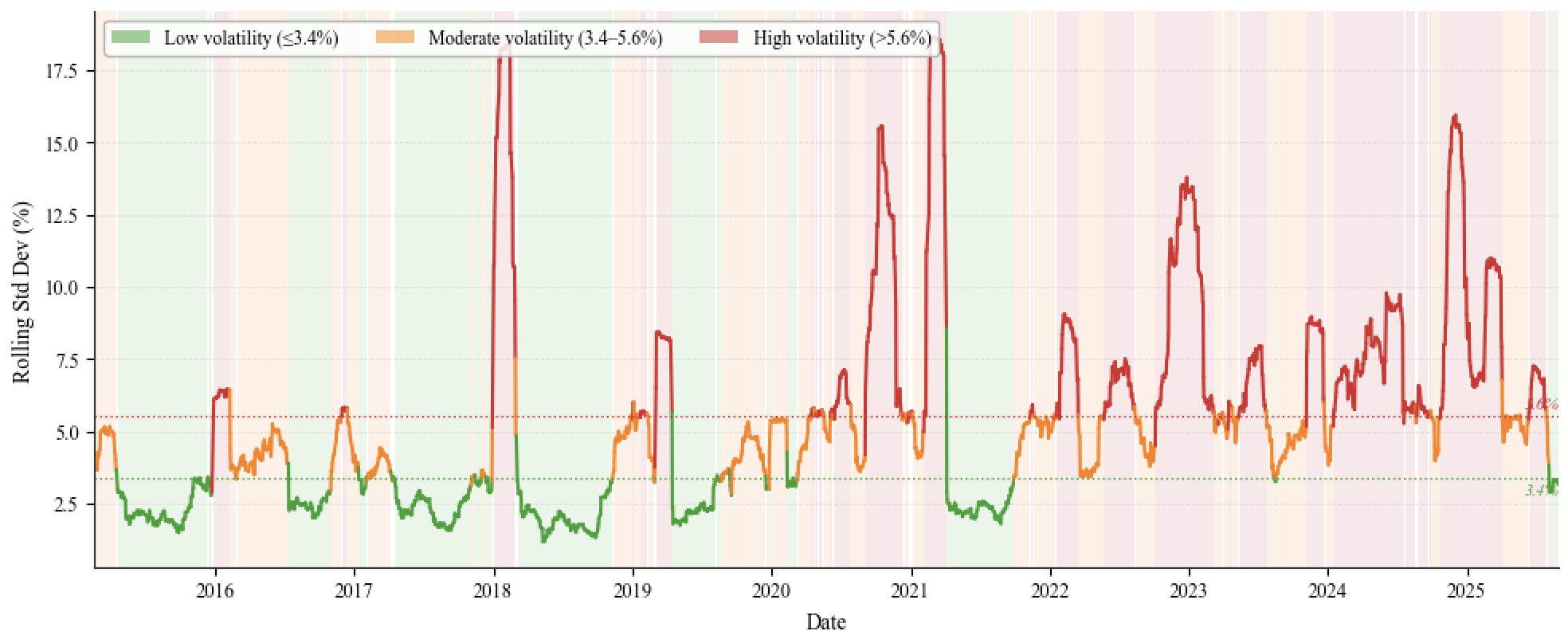}
        \caption{30-Day Rolling Standard Deviation of Spot Returns}
        \label{fig:rolling_volatility}
    \end{minipage}
\end{figure}

Figure~\ref{fig:rolling_volatility} illustrates the realized volatility of natural gas spot returns, calculated as the 30-day rolling standard deviation. The plot demonstrates distinct volatility clustering, a phenomenon where periods of relative stability are interrupted by sudden, sustained regimes of high variance. This behavior provides strong evidence of non-stationarity in the data's second moment (variance). Unlike a stationary process, where volatility would oscillate around a constant mean, the spot prices exhibit heteroskedasticity—meaning the dispersion of returns is time-dependent. These structural shifts in market risk confirm that the underlying data distribution changes over time, so it's better to use models that can adapt to evolving volatility environments.

\subsubsection{Autocorrelation Function}
\begin{figure}[H]
    \begin{minipage}[t]{0.5\textwidth}
        \centering
        \includegraphics[height=3cm]{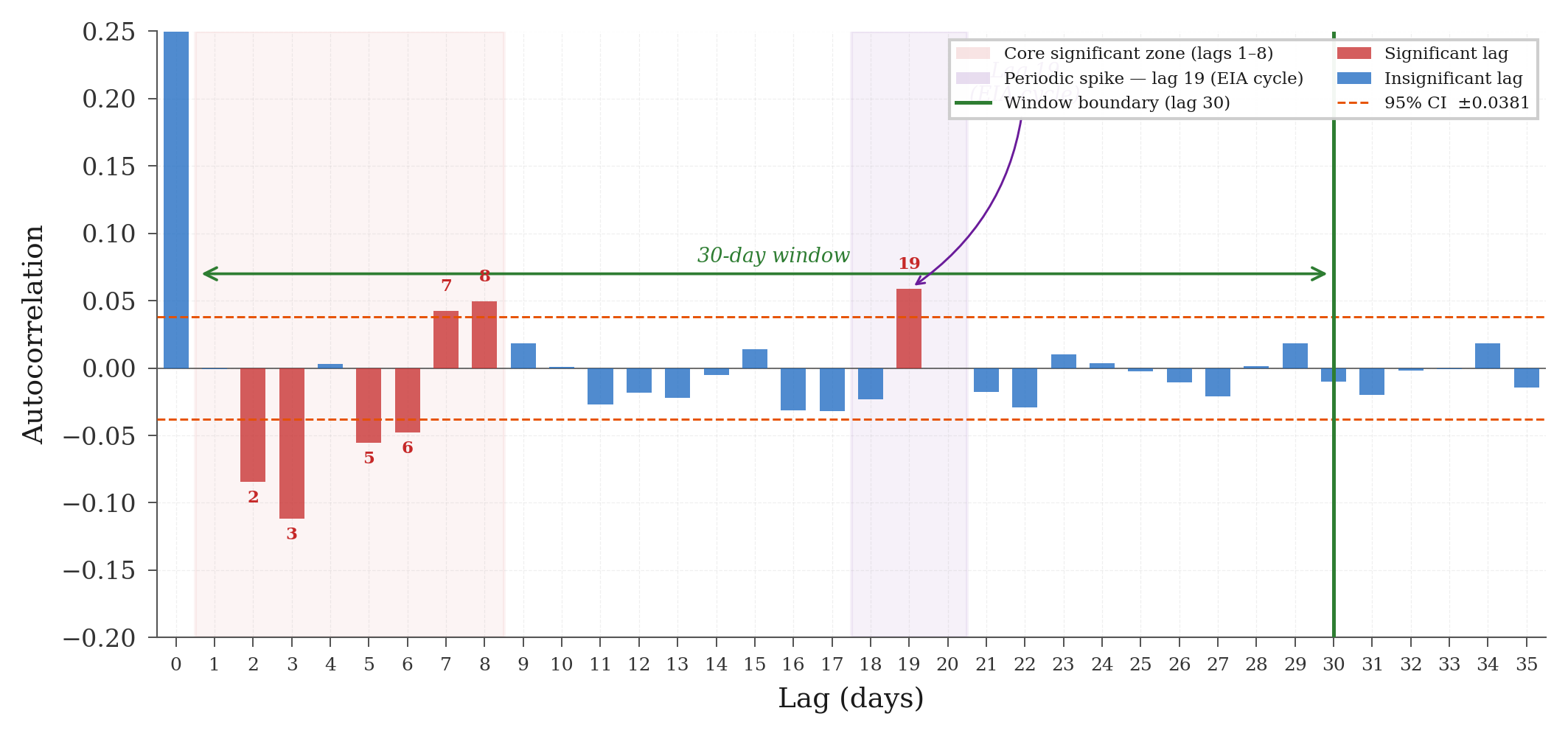}
        \caption{Autocorrelation Function of Natural Gas Spot Price Returns}
        \label{fig:acf}
    \end{minipage}
\end{figure}

The ACF plot (Figure~\ref{fig:acf}) demonstrates the Autocorrelation Function (ACF) of daily natural gas spot returns, supporting a 30-day expanding window size for the LNN model. The ACF reveals two significant autocorrelation zones relative to the 95\% confidence interval ($\pm 0.038$). The first zone, spanning lags 1--8, has the most negative autocorrelation at lag 3 (ACF $= -0.112$), indicating a short-term mean-reversion dynamic in which large price movements are partially reversed over subsequent trading days. The second zone has a statistically significant spike at lag 19 (ACF $= 0.059$) due to the monthly United States Energy Information Administration (EIA) natural gas storage reports, a market-moving event every 20 trading days. These findings led to an expanding window of 30 trading days to capture the full autocorrelation structure in the model's input sequence. 

\FloatBarrier
Collectively, exploratory analyses show that the Henry Hub natural gas price series has strong nonstationarity, volatility clustering, weak long-horizon linear dependence, and complex cross-market relationships. Based on these findings, the next section introduces this study's forecasting framework and modeling methods.

\section{Methodology}

\subsection{Baseline Model: Multivariate Rolling-window Linear Regression}

As a benchmark, we employ a multivariate rolling-window linear regression to
predict the daily return of the Henry Hub natural gas spot price. Let the spot
price at day $t$ be $P_t$. The spot return (target variable) is defined as
\begin{equation}
    R_t = \frac{P_t - P_{t-1}}{P_{t-1}}\times 100,
\end{equation}
so that $R_t$ represents the daily percentage change in the spot price.

Denote by $\mathbf{Z}_{t-1}$ the vector of exogenous predictors observed at
time $t-1$, including lagged returns and other market variables (spreads,
fundamentals, macro indicators, etc.). The baseline model at time $t$ is:
\begin{equation}
    R_t \;=\; \alpha + \beta_1 R_{t-1}
    + \boldsymbol{\gamma}^\top \mathbf{Z}_{t-1} + \varepsilon_t,
\end{equation}
where $\alpha$ is the intercept, $\beta_1$ is the autoregressive coefficient
on the lagged return, $\boldsymbol{\gamma}$ are coefficients on the exogenous
features, and $\varepsilon_t$ is the error term.

To allow for time-varying relationships, the regression is estimated in a
daily rolling window of length $W = 30$. For each forecast time $t \ge W$,
the model uses the most recent 30 observations $(t-W, \ldots, t-1)$ as the
training set. The fitted coefficients are then used to generate the
one-step-ahead forecast $\hat{R}_t$, producing a series of rolling
one-day-ahead return forecasts $\{\hat{R}_t\}$ that serve as a linear
baseline for comparison with the LNN models.

\subsection{Neural Forecasting Models}

We evaluate five neural architectures:
(i)~\textbf{LSTM},
(ii)~\textbf{Strict CfC},
(iii)~\textbf{LTC} (uniform-step continuous-time),
(iv)~\textbf{Hybrid CfC},
and (v)~\textbf{CT-LTC} (calendar-gap continuous-time).

All five models share the same input interface. Let
$\mathbf{x}_t \in \mathbb{R}^d$ denote the standardised feature vector at
time $t$. At each evaluation time $t_{\text{eval}}$, every model receives the
$L = 30$ most recent observations as a fixed-length input window:
\begin{equation}
    \mathbf{X} =
    \bigl(\mathbf{x}_{t_{\text{eval}}-L+1},
    \ldots,
    \mathbf{x}_{t_{\text{eval}}}\bigr)
    \;\in\; \mathbb{R}^{L \times d}, \qquad L = 30,
    \label{eq:nn_input}
\end{equation}
and produces a scalar one-step-ahead return forecast via a linear readout
$\hat{R}_{t_{\text{eval}}+1} = \mathbf{w}^\top \mathbf{h}_L + b$.
The window length $L$ is fixed; the \emph{training history} expands over
evaluation times (Section~\ref{sec:workflow_eval}).

\subsubsection{LSTM}
As a nonlinear recurrent benchmark, we use a standard single-layer LSTM.  
Let $\mathbf{h}_t$ and $\mathbf{c}_t$ denote the hidden and cell states at
time $t$. The recurrence is
\begin{align}
\mathbf{f}_t &= \sigma(W_f \mathbf{x}_t + U_f \mathbf{h}_{t-1} + \mathbf{b}_f), \\
\mathbf{i}_t &= \sigma(W_i \mathbf{x}_t + U_i \mathbf{h}_{t-1} + \mathbf{b}_i), \\
\mathbf{o}_t &= \sigma(W_o \mathbf{x}_t + U_o \mathbf{h}_{t-1} + \mathbf{b}_o), \\
\tilde{\mathbf{c}}_t &= \tanh(W_c \mathbf{x}_t + U_c \mathbf{h}_{t-1} + \mathbf{b}_c), \\
\mathbf{c}_t &= \mathbf{f}_t \odot \mathbf{c}_{t-1}
              + \mathbf{i}_t \odot \tilde{\mathbf{c}}_t, \\
\mathbf{h}_t &= \mathbf{o}_t \odot \tanh(\mathbf{c}_t).
\end{align}
The final hidden state is mapped to a one-step-ahead forecast by
\begin{equation}
\hat{R}_{t+1} = \mathbf{w}^{\top}\mathbf{h}_L + b.
\end{equation}
This model provides a discrete-time gated recurrent baseline without
continuous-time dynamics or liquid time constants.
\subsubsection{Strict CfC}
The Strict CfC implements the gated-interpolation form of the Closed-form
Continuous-time Neural Network~\cite{hasani2022}. Let
$\mathbf{u}_t = [\mathbf{x}_t;\,\mathbf{h}_{t-1}]$ denote the concatenation
of the current input and previous hidden state. A shared backbone MLP, given
by one linear layer followed by Tanh, maps $\mathbf{u}_t$ to
\begin{equation}
    \mathbf{s}_t
    =
    \tanh\!\left(
        W_{\mathrm{bb}}\mathbf{u}_t + \mathbf{b}_{\mathrm{bb}}
    \right)
    \in \mathbb{R}^m,
\end{equation}
where $m = \max(2n, 32)$; in our experiments, this gives $m=32$ for all
hidden sizes considered. Three independent heads then branch from
$\mathbf{s}_t$. Two Tanh heads define the candidate trajectory limits,
\begin{equation}
    \mathbf{g}_t = \tanh(W_g\mathbf{s}_t + \mathbf{b}_g),
    \qquad
    \mathbf{f}_t = \tanh(W_f\mathbf{s}_t + \mathbf{b}_f),
\end{equation}
while two linear heads parameterize the sigmoid interpolation gate,
\begin{equation}
    \mathbf{t}^{(a)}_t = W_a\mathbf{s}_t + \mathbf{b}_a,
    \qquad
    \mathbf{t}^{(b)}_t = W_b\mathbf{s}_t + \mathbf{b}_b,
\end{equation}
so that
\begin{equation}
    \boldsymbol{\sigma}_t
    =
    \sigma\!\left(
        \mathbf{t}^{(a)}_t \cdot \Delta t + \mathbf{t}^{(b)}_t
    \right),
    \qquad \Delta t = 1.
    \label{eq:strict_cfc_gate}
\end{equation}
The hidden state is updated by gated interpolation between these two
trajectory limits:
\begin{equation}
    \mathbf{h}_t
    =
    \mathbf{g}_t \odot (1 - \boldsymbol{\sigma}_t)
    +
    \boldsymbol{\sigma}_t \odot \mathbf{f}_t.
    \label{eq:strict_cfc_update}
\end{equation}
Thus, the Strict CfC replaces the explicit exponential decay of the LTC
closed-form solution with an input-dependent sigmoid interpolation gate,
while retaining a closed-form recurrent update without numerical ODE
integration. Matrix-valued trainable parameters are initialized with Xavier
uniform initialization.

\subsubsection{LTC (uniform-step)}
\label{sec:ltc}
The LTC model implements a Liquid Time-Constant network whose hidden state
evolves according to the continuous-time dynamics of
Hasani \emph{et al.}~\cite{hasani2021ltc}. For hidden unit $i$, the
underlying continuous-time dynamics are
\begin{equation}
    \frac{d x_i(t)}{dt}
    =
    -\left[
        \frac{1}{\tau_i}
        +
        f_i(\mathbf{x}(t), \mathbf{I}(t), \theta)
    \right] x_i(t)
    +
    f_i(\mathbf{x}(t), \mathbf{I}(t), \theta)\,A_i,
\end{equation}
where $\tau_i > 0$ is a learnable base time constant and $A_i$ is a
learnable attractor parameter. In the implementation, the time constants
are parameterized in log-space as
\begin{equation}
    \tau_i = \exp(\theta_{\tau,i}),
\end{equation}
which guarantees positivity.

Rather than numerically solving the ODE with a generic solver, each discrete
time step is advanced using the fused semi-implicit Euler update with
$L_{\mathrm{ode}}=6$ internal unfolding steps. Let
$\mathbf{h}^{(0)}=\mathbf{h}_{t-1}$ and
\begin{equation}
    \delta t = \frac{1}{L_{\mathrm{ode}}}.
\end{equation}
For each sub-step $\ell = 1,\dots,L_{\mathrm{ode}}$, the nonlinear response
is computed as
\begin{equation}
    \mathbf{f}^{(\ell)}
    =
    \sigma\!\left(
        W_{\mathrm{rec}}\mathbf{h}^{(\ell-1)}
        +
        W_{\mathrm{in}}\mathbf{x}_t
        +
        \mathbf{b}
    \right),
\end{equation}
where $W_{\mathrm{in}}$ is an input-to-hidden linear map computed once per
external time step and reused across all sub-steps, $W_{\mathrm{rec}}$ is a
recurrent hidden-to-hidden linear map recomputed at each sub-step, and
$\mathbf{b}$ is a learnable bias vector. The hidden state is then updated by
\begin{equation}
    \mathbf{h}^{(\ell)}
    =
    \frac{
        \mathbf{h}^{(\ell-1)}
        +
        \delta t\,\mathbf{f}^{(\ell)} \odot \mathbf{A}
    }{
        \mathbf{1}
        +
        \delta t\left(
            \mathbf{1}/\boldsymbol{\tau}
            +
            \mathbf{f}^{(\ell)}
        \right)
    },
    \label{eq:ltc_fused_update}
\end{equation}
and the final hidden state for the external time step is
\begin{equation}
    \mathbf{h}_t = \mathbf{h}^{(L_{\mathrm{ode}})}.
\end{equation}

\subsubsection{Hybrid CfC}
The Hybrid CfC uses a closed-form recurrent update with input-dependent
time-scale modulation. At each step, the input and previous hidden state are
concatenated as
\begin{equation}
    \mathbf{u}_t = [\mathbf{x}_t;\,\mathbf{h}_{t-1}].
\end{equation}
A nonlinear drive and adaptive time constant are then computed by
\begin{align}
    \mathbf{f}_t &= \tanh(W_f \mathbf{u}_t + \mathbf{b}_f), \\
    \boldsymbol{\tau}_t &=
    \exp(\boldsymbol{\theta}_\tau)\odot
    \sigma(W_\tau \mathbf{u}_t + \mathbf{b}_\tau),
\end{align}
with elementwise lower clipping
\begin{equation}
    \boldsymbol{\tau}_t \leftarrow \max(\boldsymbol{\tau}_t,\,10^{-3}).
\end{equation}
The effective decay and exponential gate are
\begin{align}
    \mathbf{A}_t &= \frac{1}{\boldsymbol{\tau}_t} + |\mathbf{f}_t|, \\
    \mathbf{g}_t &= \exp\!\bigl(-\max(\mathbf{A}_t,\,10^{-3})\bigr).
\end{align}
The hidden state update is
\begin{equation}
    \mathbf{h}_t
    =
    \mathbf{g}_t \odot \mathbf{h}_{t-1}
    +
    (1-\mathbf{g}_t)\odot \frac{\mathbf{f}_t}{\mathbf{A}_t},
    \label{eq:hybrid_cfc_update}
\end{equation}
which can be interpreted as exponential relaxation toward the
data-dependent equilibrium $\mathbf{f}_t/\mathbf{A}_t$.
\subsubsection{CT-LTC (calendar $\Delta t$)}
CT-LTC extends the LTC formulation (Section~\ref{sec:ltc}) by replacing the
fixed unit step size with the observed calendar gap between consecutive
timestamps. The underlying ODE, log-space time constant parameterization,
nonlinear response, and fused semi-implicit Euler update structure are
identical to those of LTC. The sole modification is that the sub-step size
is determined by the elapsed calendar time $\Delta_t$ between observations
at times $t-1$ and $t$, measured in days and clipped below at 1:
\begin{equation}
    \delta t_t = \frac{\Delta_t}{L_{\mathrm{ode}}}.
\end{equation}
The hidden state update at each internal sub-step then follows
Eq.~(\ref{eq:ltc_fused_update}) with $\delta t$ replaced by $\delta t_t$.
Weekend and holiday gaps therefore induce larger integration steps than
ordinary adjacent trading days, making CT-LTC sensitive to irregular
calendar spacing in the observed financial series.

\subsection{Workflow and Evaluation Protocol}
\label{sec:workflow_eval}

\subsubsection*{Phase 0: Data Preparation}

The full dataset of $N$ observations is partitioned chronologically at the
50\% mark into a \emph{tuning set} (first $\lfloor N/2 \rfloor$ rows) and an
\emph{evaluation set} (remaining rows). All hyperparameter search is conducted
exclusively on the tuning set; the evaluation set is never observed during
tuning.

Feature exclusion is consistent across all models, with one exception. The
non-CT runs exclude \{Date, Spot Return, Spot Price, AR1 Spot Price\},
leaving 30 predictor features. The CT-LTC run additionally excludes the
lagged return to avoid direct target leakage, leaving 29 predictor features.

\subsubsection*{Phase 1: Hyperparameter Tuning}

Within the tuning set, a time-ordered 80/20 split yields an inner training
segment and an inner validation segment. A \texttt{StandardScaler} is fitted
on the inner training segment only and applied to both; no validation data
influence the scaler. For each candidate configuration, the model is trained
for 30 epochs using the Adam optimiser with weight decay $\lambda = 10^{-5}$
and gradient clipping at max $\ell_2$-norm $= 1.0$. Mean squared error (MSE)
is used as the training loss. The configuration achieving the highest Pearson
correlation $r$ on the inner validation segment is selected.

All five neural models share the hyperparameter grid shown in
Table~\ref{tab:hyperparam_shared}.
\vspace{-4pt}

\begin{table}[htbp]
\centering
\caption{Hyperparameter grid shared by all five neural models.}
\label{tab:hyperparam_shared}
\begin{tabular}{lc}
\toprule
Hyperparameter  & Candidate values \\
\midrule
Hidden size      & $\{4,\;8,\;12\}$               \\
Learning rate    & $\{5\times10^{-3},\;10^{-3}\}$ \\
Batch size       & $\{32,\;64\}$                  \\
Number of layers & $\{1\}$                        \\
\bottomrule
\end{tabular}

\vspace{4pt}
{\footnotesize\textit{Note:} LTC and CT-LTC additionally fix
$L_{\mathrm{ode}} = 6$ ODE sub-steps (not searched).}
\end{table}

The search space is intentionally compact for two reasons. First, LNN
architectures are parameter-efficient by design: continuous-time dynamics
and input-dependent time constants encode rich temporal structure with far
fewer hidden units than comparably expressive discrete recurrent
networks~\cite{hasani2021ltc,hasani2022}, making hidden sizes of 4, 8, and 12
sufficient for a daily financial series of roughly 2,600 observations.
Second, the expanding-window protocol requires retraining from scratch at
each of the $N_{\text{eval}} = 160$ evaluation points; the full grid already
exceeds 9,000 individual training runs across all five architectures, and
extending it further would increase cost superlinearly with diminishing
benefit. Fixing the number of layers at one is consistent with empirical
findings showing that single-layer LTC and CfC cells match or exceed deeper
discrete recurrent stacks on short-horizon regression tasks~\cite{hasani2021ltc}.

\subsubsection*{Phase 2: Stratified Expanding-Window Evaluation}
Out-of-sample performance is assessed using a deterministic stratified
expanding-window scheme. An expanding rather than rolling window is used
because LNNs benefit from growing training sets; discarding historical data
would systematically disadvantage them relative to the linear baseline.
The evaluation indices are divided into $K = 20$ equal-length temporal bins
with $k = 8$ evenly spaced points drawn from each, yielding $N_{\text{eval}}
= 160$ evaluation points applied identically across all architectures. At
each $t_{\text{eval}}$, all preceding observations form the expanding
training set; a \texttt{StandardScaler} is fitted exclusively on this
window; the model is retrained from scratch for 50 epochs under the tuned
hyperparameters; and the one-step-ahead forecast is generated from the
$L = 30$ most recent standardised observations.
\subsubsection*{Phase 2b: Moving Block Bootstrap}

To quantify sampling uncertainty of out-of-sample metrics, we apply a Moving
Block Bootstrap (MBB) to the $N_{\text{eval}}$ stored forecast--truth pairs
$\{(y_i, \hat{y}_i)\}_{i=1}^{N_{\text{eval}}}$, with all model predictions
held fixed. Each bootstrap replication draws a resample of size
$N_{\text{eval}}$ by selecting random starting indices (with wrap-around) and
concatenating contiguous blocks of length $\ell_{\text{block}}$, then
recomputes all performance metrics on the resampled pairs. MBB preserves the
temporal autocorrelation of the forecast error series that i.i.d.\ resampling
would destroy.

The block length is chosen data-adaptively from the ACF of residuals
$e_i = y_i - \hat{y}_i$:
\begin{equation}
    \ell_{\text{block}} \;=\;
    \max\!\left(5,\;\min\!\left(
        \hat{\ell}_{\mathrm{ACF}} + 2,\;
        \left\lfloor \frac{N_{\text{eval}}}{10} \right\rfloor
    \right)\right),
\end{equation}
where $\hat{\ell}_{\mathrm{ACF}}$ is the last lag at which
$|\hat{\rho}(\ell)| > 1/\sqrt{N_{\text{eval}}}$ in the sample ACF. We use
$B = 300$ bootstrap replications and report the bootstrap mean, standard
deviation, and 2.5th--97.5th percentile interval (95\% CI) for each metric.

\subsubsection*{Training Details}

All five neural models share the same settings: sequence length $L=30$ (fixed lookback window), 30 epochs per tuning trial, 50 epochs per expanding-window retrain, and silent skipping of batches containing NaN inputs, predictions, or losses.
\subsubsection*{Selection Criterion}
The primary model selection metric is the Pearson correlation $r$ between predicted and actual returns on the validation segment. The hyperparameter set with the highest validation Pearson $r$ is retained as the best configuration.

\section{Results}
\begin{table*}[!htbp]
\centering
\caption{Model performance on the test set and under bootstrap resampling.}
\label{tab:results}
\setlength{\tabcolsep}{5pt}
\renewcommand{\arraystretch}{1.05}
\begin{tabular}{lcccccc}
\toprule
\multicolumn{7}{c}{\textbf{Panel A: Test-set performance}} \\
\midrule
\textbf{Model} &
\textbf{Pearson $r$} &
\textbf{Spearman $\rho$} &
\textbf{DA (\%)} &
\textbf{$R^2$} &
\textbf{RMSE} &
\textbf{MAE} \\
\midrule
Baseline   & $-0.0408$ & $-0.0122$ & $43.30$ & $-91.03$ & $61.12$ & $32.45$ \\
LSTM       & $0.1035$  & $0.1714$  & $55.00$ & $-0.0133$ & $8.129$ & $5.149$ \\
Strict CfC & $0.1059$  & $0.1593$  & $52.50$ & $-0.0528$ & $8.285$ & $5.398$ \\
LTC        & $0.2548$  & $0.2262$  & $51.88$ & $0.0540$  & $7.854$ & $5.026$ \\
Hybrid CfC & $0.3011$  & $0.2680$  & $53.75$ & $0.0573$  & $7.840$ & $4.968$ \\
CT-LTC     & $0.0972$  & $0.1307$  & $52.50$ & $0.0076$  & $8.044$ & $5.087$ \\
\midrule
\multicolumn{7}{c}{\textbf{Panel B: Bootstrap performance (mean $\pm$ std, 95\% CI $[2.5^{\text{th}},\,97.5^{\text{th}}]$, $R = 300$)}} \\
\midrule
\textbf{Model} &
\textbf{Pearson $r$} &
\textbf{Spearman $\rho$} &
\textbf{DA (\%)} &
\textbf{$R^2$} &
\textbf{RMSE} &
\textbf{MAE} \\
\midrule
LSTM
  & $0.119 \pm 0.099$     & $0.164 \pm 0.055$     & $55.08 \pm 2.67$     & $-0.009 \pm 0.039$    & $8.014 \pm 1.800$     & $5.168 \pm 0.740$ \\
  & $[-0.058,\ 0.294]$    & $[0.063,\ 0.272]$     & $[50.63,\ 60.00]$    & $[-0.064,\ 0.081]$    & $[5.271,\ 11.502]$    & $[3.972,\ 6.747]$ \\[3pt]
Strict CfC
  & $0.110 \pm 0.055$     & $0.150 \pm 0.060$     & $52.42 \pm 3.35$     & $-0.069 \pm 0.042$    & $8.207 \pm 1.656$     & $5.421 \pm 0.753$ \\
  & $[0.016,\ 0.227]$     & $[0.039,\ 0.283]$     & $[46.25,\ 59.38]$    & $[-0.164,\ -0.004]$   & $[5.661,\ 11.455]$    & $[4.148,\ 7.116]$ \\[3pt]
LTC
  & $0.259 \pm 0.054$     & $0.222 \pm 0.050$     & $51.92 \pm 2.20$     & $0.049 \pm 0.016$     & $7.761 \pm 1.651$     & $5.041 \pm 0.714$ \\
  & $[0.163,\ 0.368]$     & $[0.128,\ 0.314]$     & $[47.50,\ 55.63]$    & $[0.018,\ 0.080]$     & $[5.281,\ 10.972]$    & $[3.923,\ 6.620]$ \\[3pt]
Hybrid CfC
  & $\mathbf{0.302 \pm 0.052}$  & $\mathbf{0.263 \pm 0.058}$  & $\mathbf{53.80 \pm 3.86}$  & $\mathbf{0.051 \pm 0.016}$  & $\mathbf{7.750 \pm 1.638}$  & $\mathbf{4.986 \pm 0.708}$ \\
  & $\mathbf{[0.194,\ 0.398]}$  & $\mathbf{[0.151,\ 0.365]}$  & $\mathbf{[45.63,\ 60.00]}$ & $\mathbf{[0.015,\ 0.084]}$  & $\mathbf{[5.202,\ 10.950]}$ & $\mathbf{[3.840,\ 6.545]}$ \\[3pt]
CT-LTC
  & $0.112 \pm 0.100$     & $0.128 \pm 0.093$     & $52.58 \pm 3.07$     & $0.006 \pm 0.019$     & $7.939 \pm 1.712$     & $5.103 \pm 0.747$ \\
  & $[-0.063,\ 0.304]$    & $[-0.060,\ 0.304]$    & $[45.63,\ 58.13]$    & $[-0.026,\ 0.047]$    & $[5.252,\ 11.293]$    & $[3.937,\ 6.732]$ \\
\bottomrule
\end{tabular}
\bigskip

\noindent\small\textit{Note:} Panel B reports bootstrap mean $\pm$ standard deviation with the 95\% confidence
interval (2.5\textsuperscript{th}--97.5\textsuperscript{th} percentile) on the second line of each model row.
Best value in each Panel~B column is \textbf{bolded}.
\end{table*}

\subsection{Metrics}

To evaluate predictive performance, we employ Pearson's correlation coefficient, Spearman's rank correlation coefficient, directional accuracy, the coefficient of determination ($R^{2}$), root mean squared error (RMSE), and mean absolute error (MAE). Pearson's correlation serves as the primary model selection and evaluation metric for two reasons. Scale-dependent metrics such as $R^{2}$, RMSE, and MAE can become unstable during structural breaks or volatility spikes, where a single extreme observation may dominate the aggregate statistic. Directional accuracy evaluates the fraction of correctly predicted return signs:

\begin{equation}
\text{DA} = \frac{1}{n} \sum_{i=1}^{n} \mathbb{I}[\text{sign}(y_i) = \text{sign}(\hat{y}_i)].
\end{equation}

Together, Pearson's $r$ captures proportional co-movement, Spearman's $\rho$ reflects monotonic structure, and DA measures practical forecasting value in trading applications where directional correctness often matters more than exact magnitude.

\subsection{Results and Analysis}

\subsubsection{Extreme Event Behavior and Forecast Deviations}
For all model predictions, visualizations comparing predicted and actual
returns reveal several large and abrupt deviations that none of the evaluated
models adequately capture, reflecting extreme market movements caused by
exogenous shocks rather than typical market dynamics.

One example occurred in February 2021 during Winter Storm Uri, when
extremely cold temperatures increased heating demand while freezing conditions
disrupted production and infrastructure, particularly in Texas, significantly
straining natural gas and electricity markets.~\cite{eia_uri_2021} A similar
pattern was observed in early 2022, when prices rose due to high storage
withdrawals, strong LNG exports, and declining production exacerbated by cold
weather.~\cite{eia_2022_winter_tightening} Another fluctuation occurs in
December 2022 during Winter Storm Elliott~\cite{eia_elliott_2022}, while the
spike around April 2024 reflects transient market imbalances such as strong
LNG export demand and thin market liquidity.~\cite{eia_2024_spring_market}
Finally, in November 2024, an abrupt reversal after exceptionally low prices
generates disproportionately large positive percentage returns as anticipated
winter risks triggered a quick recovery.~\cite{eia_2024_storage_high}

These findings highlight a key limitation of data-driven models: even
adaptive architectures cannot anticipate exogenous shocks outside the
historical data distribution, suggesting that incorporating real-time
external signals such as weather forecasts may be required to better capture
tail-risk events.
\subsubsection{Baseline Model: AR(1) Spot Price + All Features}
\begin{figure}[H]
    \centering
    \includegraphics[width=0.9\columnwidth]{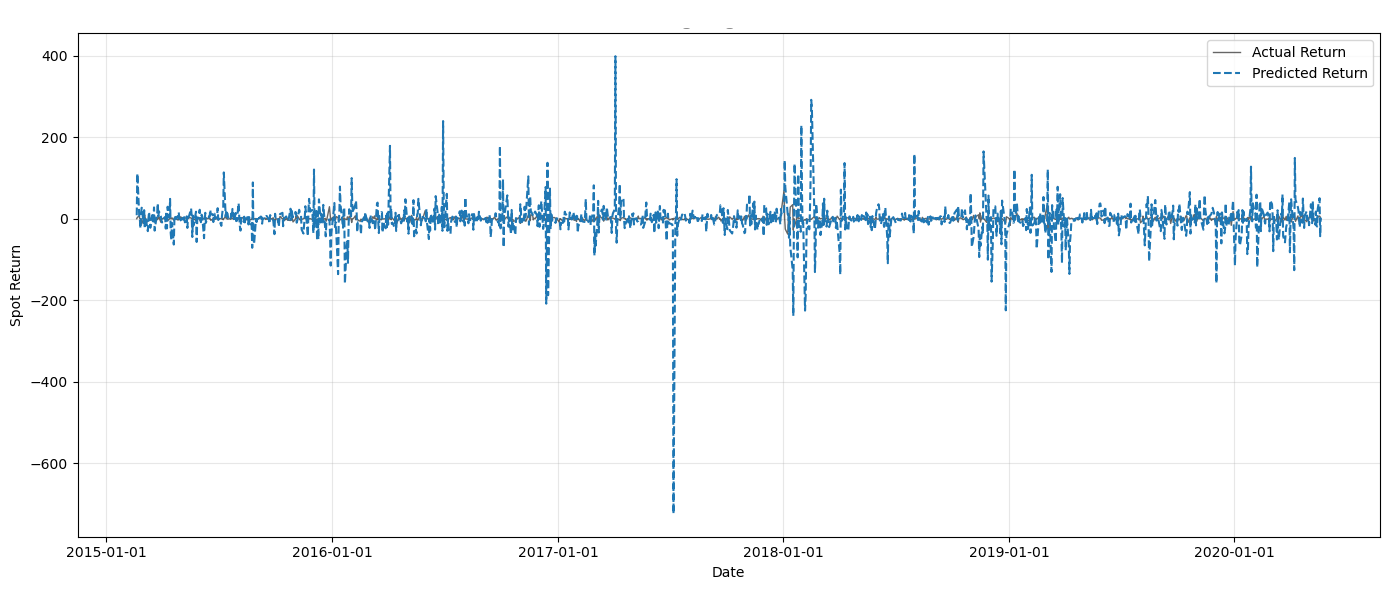}
    \caption{Baseline 30-Day Rolling Regression: Actual vs.\ Predicted Spot Returns (First 50\% of Timeline)}
    \label{fig:Baseline_Results_Part1}
    
    \vspace{-0.3cm}
    
    \includegraphics[width=0.9\columnwidth]{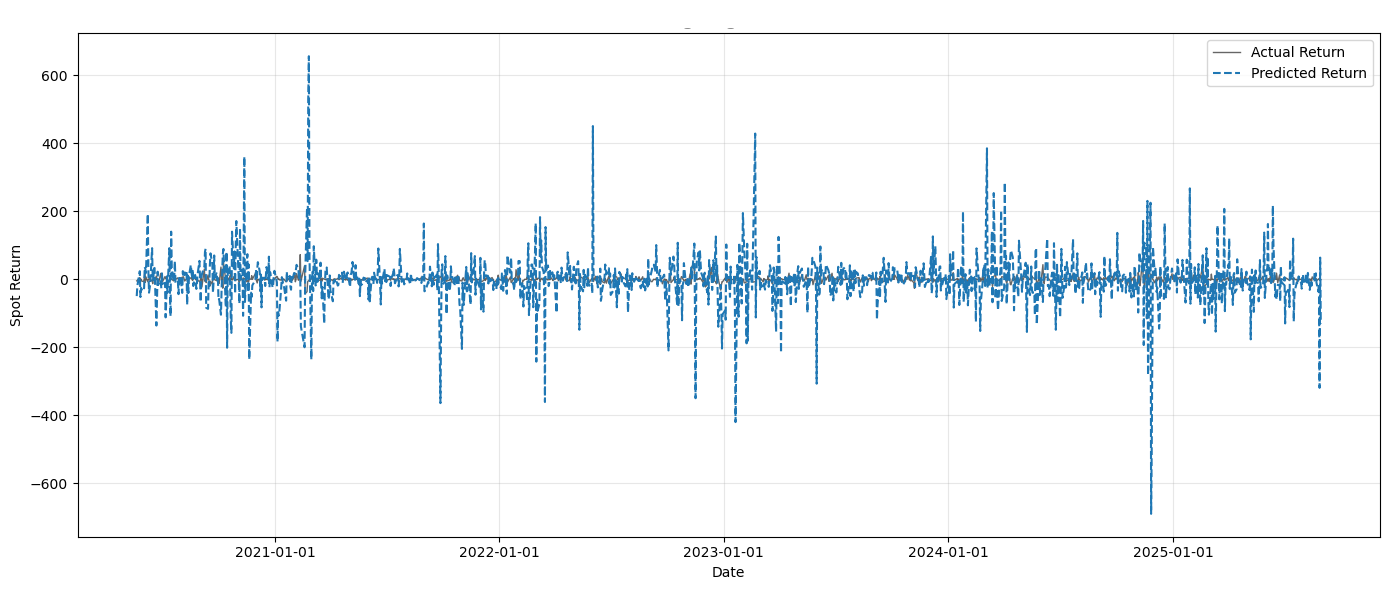}
    \caption{Baseline 30-Day Rolling Regression: Actual vs.\ Predicted Spot Returns (Second 50\% of Timeline)}
    \label{fig:Baseline_Results_Part2}
\end{figure}
Using a 30-day rolling window, the baseline model generated 2,615 one-step-ahead forecasts for daily spot price percentage changes. Correlation-based metrics indicated no predictive structure, with Pearson's $r = -0.039$ and Spearman's $\rho = -0.012$, and directional accuracy of 43.33\% failed to outperform a random 50\% benchmark. Error metrics were substantial (RMSE: 61.12; MAE: 32.45). The $R^{2}$ of $-91.03$ indicates that the baseline model's forecasts explain far less variance than a naive mean predictor; while such extreme negative values may appear unusual, they are a common outcome when a linear model with many parameters is applied to heavy-tailed, regime-switching return series where the conditional mean is highly unstable, causing the sum of squared forecast errors to greatly exceed the total variance of the target. This result confirms the fundamental limitations of linear models in this forecasting setting.

\subsubsection{LSTM}
As reported in Table~\ref{tab:results}, the LSTM achieves the highest
directional accuracy among the non-hybrid architectures, yet its $R^2$
remains near zero and its correlation coefficients are the weakest in the
neural model tier. This dissociation reflects a structural limitation of
the discrete gating mechanism: the forget, input, and output gates regulate
information flow through fixed sigmoid activations that are insensitive to
both the elapsed time between observations and the magnitude of the current
input shock. Consequently, while the model can associate input
configurations with a return direction, it cannot modulate the rate at
which stale hidden-state information is discarded in response to abrupt
market events. In a series characterised by sudden regime shifts,
volatility clustering, and heavy-tailed spikes, this rigidity causes the
hidden state to retain outdated information and smooth through the dynamics
that drive magnitude variation.

Bootstrap resampling confirms that this weakness is a systematic property
of the architecture rather than an artefact of the evaluation window. The
confidence intervals for both Pearson $r$ and $R^2$ straddle zero,
indicating that co-movement with realised returns and explanatory power
relative to a naive mean predictor are both unreliable across different
market regimes. The LSTM therefore serves as a useful nonlinear recurrent
benchmark but is ill-suited to the adaptive demands of volatile commodity
forecasting.
\FloatBarrier
\begin{figure}[!htbp]
    \centering
    \includegraphics[width=0.9\columnwidth]{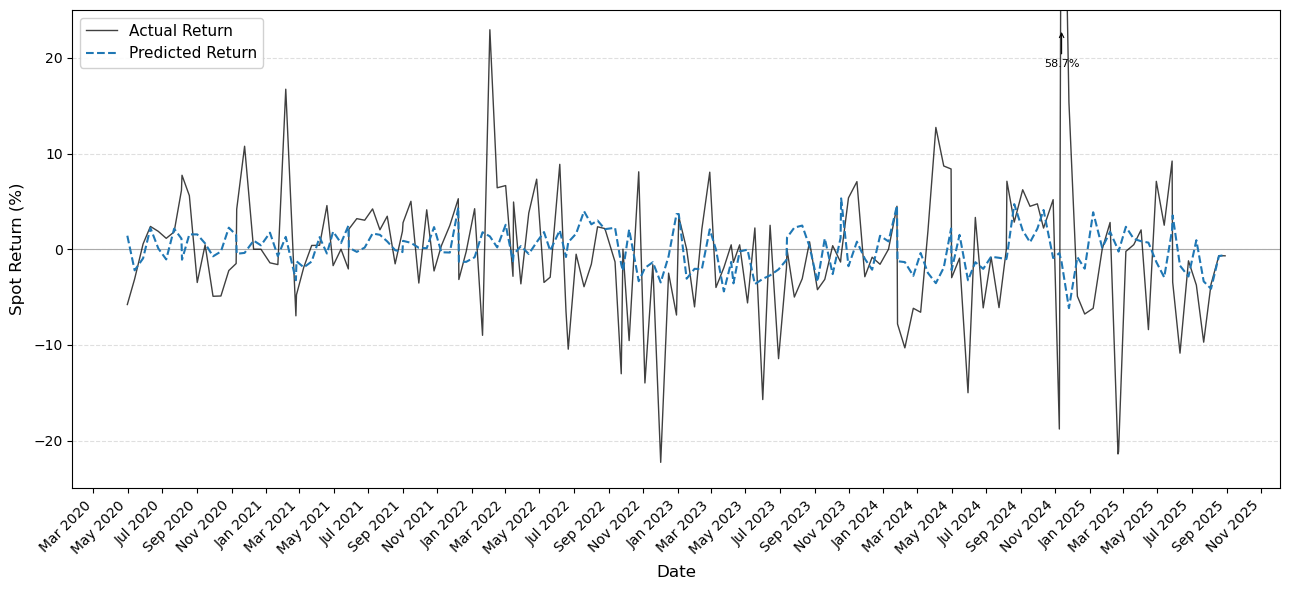}
    \caption{LSTM model: actual vs.\ predicted spot returns.}
    \label{fig:LSTM_Results}
\end{figure}
\FloatBarrier

\subsubsection{Strict CfC}
As shown in Table~\ref{tab:results}, the Strict CfC fails to outperform
the LSTM in any metric of practical significance. Despite its
continuous-time motivation, the architecture's sigmoid interpolation gate
controls only the mixture ratio between two candidate trajectory limits,
not the rate at which the hidden state transitions between them. A large
return shock can shift the blend toward a new candidate state, but the
transition carries no explicit timescale dependence, leaving the model
with an inertia that is unsuitable for a series in which consecutive
observations can differ by an order of magnitude.
The $R^2$ bootstrap distribution demonstrates structural rigidity, with a confidence interval below 0. Sample mean return forecasts explain more than strict CfC forecasts across all resampled evaluation windows. Architectures that capture continuous-time dynamics are especially affected. The model's limited positive co-movement in point-estimate evaluation may be due to favorable sampling, not a strong learned signal, as both correlation coefficients have wide confidence intervals. As the weakest neural architecture, strict CfC shows that the closed-form continuous-time formulation without input-conditioned timescale modulation does not outperform a standard recurrent benchmark in this forecasting scenario.
\FloatBarrier
\begin{figure}[!htbp]
    \centering
    \includegraphics[width=0.9\columnwidth]{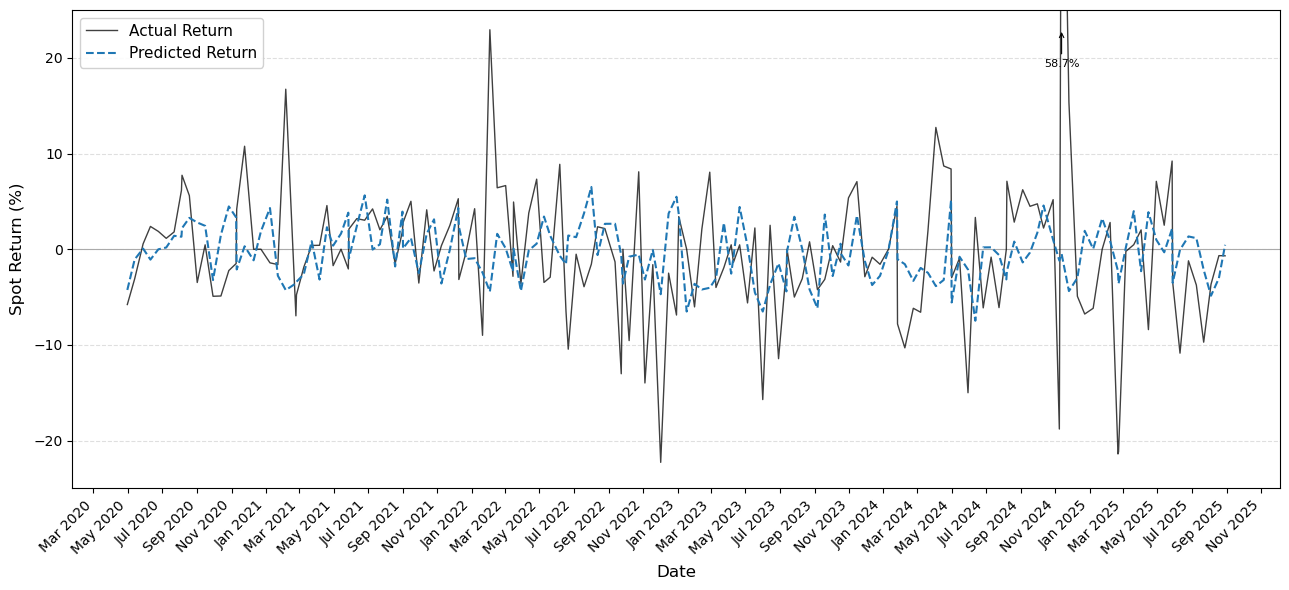}
    \caption{Strict CfC model: actual vs.\ predicted spot returns.}
    \label{fig:Strict_CfC_Results}
\end{figure}
\FloatBarrier

\subsubsection{LTC}
The LTC model greatly outperforms the LSTM and Strict CfC, as shown in Table~\ref{tab:results}.
Pearson and Spearman correlations are nearly double those of weaker architectures, with positive $R^2$ and low error metrics in the neural tier. Significantly, these improvements focus on magnitude-related metrics, while directional accuracy remains modest. The LTC architecture's continuous-time ODE formulation, governed by learnable time constants and attractor parameters, enables the model to track return path levels and shapes more accurately than discrete recurrent networks.
The LTC solves hidden-state dynamics with multiple internal sub-steps per observation, allowing for finer temporal resolution than the observed daily frequency. This allows for smooth drift components to be captured, even with noisy surface-level returns.
\FloatBarrier
\begin{figure}[!htbp]
    \centering
    \includegraphics[width=0.9\columnwidth]{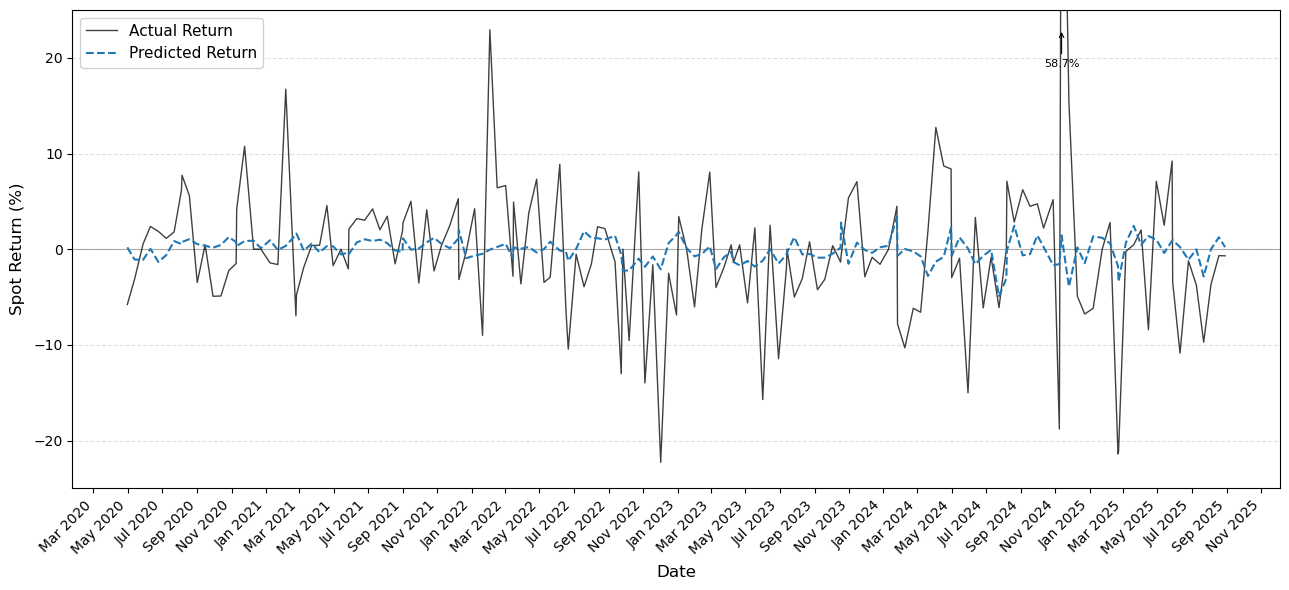}
    \caption{Liquid Time-Constant (LTC) model: actual vs.\ predicted spot
    returns.}
    \label{fig:LTC_LNN_Results}
\end{figure}
\FloatBarrier
The Boostrap results show that the Pearson $r$ and $R^2$ confidence intervals are strictly positive across the bootstrap distribution. This property is only shared by Hybrid CfC among all evaluated models. The advantage of LTC over a naive predictor is not due to a favorable evaluation window, but rather to a reliable and consistent learned signal. The LTC's uniform sub-step size $\Delta t = 1/L_{\text{ode}}$, which remains constant regardless of input dynamics, is an important architectural constraint. As shown in Figure 5, the model's ability to represent both short-horizon mean-reversion dynamics and longer-horizon trend components is limited by the fact that all hidden units integrate at the same frequency. The Hybrid CfC addresses this constraint, explaining why, despite its high performance, LTC is consistently outperformed by its hybrid counterpart.

\subsubsection{Hybrid CfC}
The Hybrid CfC is the best overall performer across all metrics in both
the point-estimate and bootstrap panels, as shown in Table~\ref{tab:results}.
Its superiority over the LTC — the second-strongest model — is narrow in
absolute terms but consistent across all resampled evaluation windows,
which is the more important criterion for assessing genuine architectural
advantage. The mechanism underlying this edge lies in the adaptive decay
rate $A_t = \frac{1}{\tau_t} + |f_t|$ (Eq.~26), which couples a learned
base time constant with the instantaneous magnitude of the driving signal.
This single modification has a profound consequence for how the model
responds to market shocks: when a large return is observed, $|f_t|$
increases, directly raising the effective decay rate and causing the hidden
state to shed stale information more rapidly. Conversely, during quiet
periods, the decay rate falls toward the baseline $1/\tau_t$, allowing the
model to integrate information over a longer effective horizon. This
dynamic adjustment is precisely what is required for a series with the
empirical properties of Henry Hub returns — abrupt, heavy-tailed shocks
that demand rapid state updating, interspersed with calmer regimes where
longer memory improves forecast stability.
\FloatBarrier
\begin{figure}[!htbp]
    \centering
    \includegraphics[width=0.9\columnwidth]{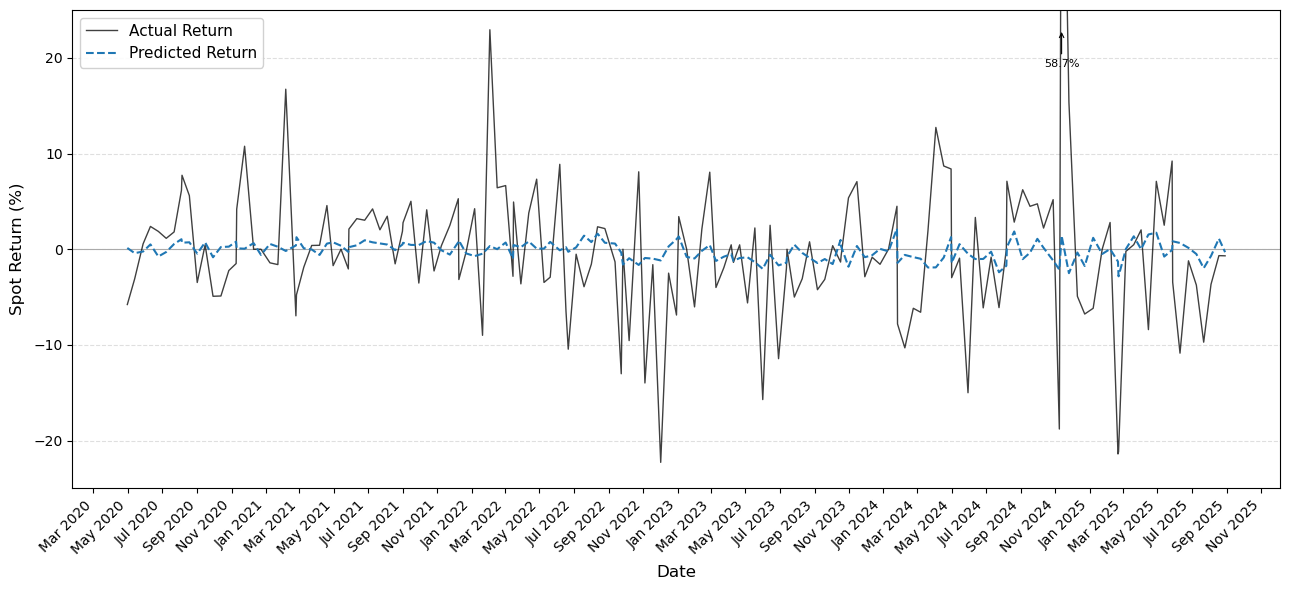}
    \caption{Hybrid CfC model: actual vs.\ predicted spot returns.}
    \label{fig:Hybrid_CfC_Results}
\end{figure}
\FloatBarrier
The bootstrap $R^2$ confidence interval lying entirely above zero provides
the strongest statistical evidence that the Hybrid CfC's advantage is
systematic rather than sample-specific. The partial overlap between its
$R^2$ interval and that of LTC suggests the two architectures occupy a
statistically close superior tier, well separated from the LSTM, Strict
CfC, and CT-LTC. The relatively wide directional accuracy interval,
however, indicates that sign prediction remains noisy — the model's
primary contribution is to return-level fidelity rather than binary
directional classification. Taken together, the results suggest that
input-conditioned timescale modulation is the single most important
architectural feature for this forecasting task, and that the Hybrid CfC's
implementation of this property via a fully closed-form update represents
the most effective balance between temporal adaptability and computational
tractability among the architectures considered.

\subsubsection{CT-LTC}
The CT-LTC model extends the LTC formulation by replacing the fixed unit
step size with the observed calendar gap between consecutive timestamps,
encoding weekend and holiday spacing directly into the integration
dynamics. Despite the theoretical appeal of this modification — financial
time series are inherently irregularly spaced, and Monday returns
aggregate three calendar days of information — the empirical results in
Table~\ref{tab:results} reveal that calendar-aware step sizing does not
translate into improved forecasting performance relative to the uniform-step
LTC. This counterintuitive finding warrants careful interpretation.
\FloatBarrier
\begin{figure}[!htbp]
    \centering
    \includegraphics[width=0.9\columnwidth]{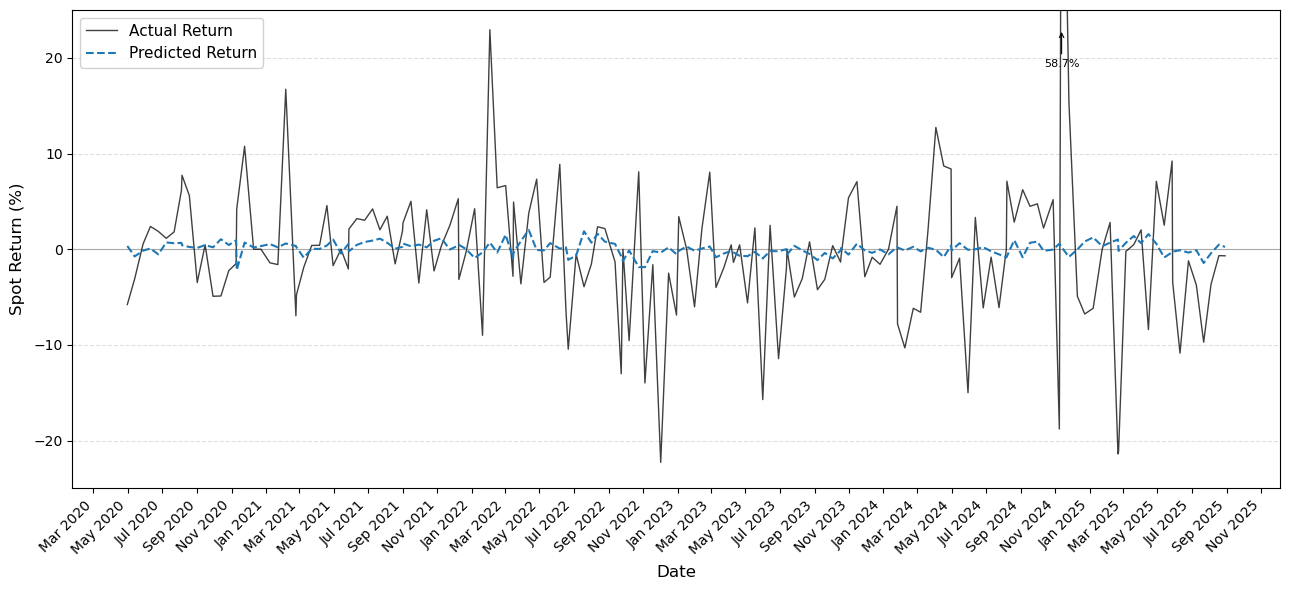}
    \caption{CT-LTC model: actual vs.\ predicted spot returns.}
    \label{fig:CTLTC_Results}
\end{figure}
\FloatBarrier

The fundamental limitation of CT-LTC is that a single calendar-derived
$\Delta t$ is applied uniformly across all hidden units, forcing every unit
to integrate at the same rate regardless of the informational content of
the current input. Calendar distance is a poor proxy for informational
distance in a market driven by discontinuous, event-specific shocks: a
Monday following a quiet weekend and one following a major supply
disruption carry very different information loads, yet receive identical
step sizes. The Hybrid CfC avoids this by conditioning its effective
timescale on $|f_t|$, tracking signal magnitude rather than elapsed time.

Bootstrap confidence intervals for both Pearson $r$ and $R^2$ straddle
zero, confirming that CT-LTC's marginal point-estimate gains are not
reproducible across evaluation windows. The architecture is a principled
extension of LTC, but its calendar-gap parameterisation imposes temporal
rigidity incompatible with the shock-driven dynamics of the Henry Hub
market. Replacing uniform calendar spacing with input-conditioned $\Delta t$
scaling represents a natural direction for future work.
\subsubsection{Forecast Bias}
 
To assess whether the evaluated models produce systematically biased forecasts, we compute the mean forecast error for each architecture and test its significance using a two-sided $t$-test. A significantly nonzero mean error would indicate that a model consistently over- or under-predicts spot returns.
 
None of the five neural architectures exhibits statistically significant forecast bias at the 5\% level. The LSTM produces a mean error of $-0.250$ ($t = -0.388$, $p = 0.699$), indicating a slight but insignificant tendency to over-predict returns. The Strict CfC is the least biased model in absolute terms, with a mean error of $+0.047$ ($t = 0.072$, $p = 0.943$). The LTC and Hybrid CfC show modest negative mean errors of $-0.213$ ($t = -0.343$, $p = 0.732$) and $-0.113$ ($t = -0.182$, $p = 0.856$), respectively. The CT-LTC exhibits the largest absolute bias at $-0.336$ ($t = -0.527$, $p = 0.599$), consistent with its tendency to under-react to positive return episodes, though this bias remains statistically insignificant.
 
\section{Discussion}
\subsection{Conclusion}

Forecasting natural gas spot returns remains inherently challenging due to the market's pronounced nonlinearity, heavy-tailed distributions, and frequent regime shifts. The Henry Hub series exhibits not only high volatility but also abrupt structural changes that violate the assumptions underlying many traditional time-series models. As a result, both linear methods and standard recurrent neural architectures often struggle to capture the true dynamics of the data-generating process.

This study compares a variety of Liquid Neural Network (LNN) architectures---including LTC, Strict Closed-form Continuous-time (CfC), Hybrid CfC, and CT-LTC---with a multivariate rolling-window linear regression baseline and a standard LSTM benchmark. The empirical results demonstrate that LNN-based models consistently outperform the baseline and offer meaningful advantages over traditional recurrent approaches, confirming that architectures with adaptive temporal dynamics are better suited to the nonlinear and nonstationary behavior observed in energy markets.

Among the evaluated models, the Hybrid CfC and LTC architectures emerge as the most effective. Both achieve higher out-of-sample $R^{2}$ values, stronger linear and rank correlations with realized returns, and lower forecast errors relative to competitors. Their advantage stems not merely from a continuous-time formulation but from the ability to balance temporal adaptability with nonlinear flexibility. The Hybrid CfC, in particular, delivers the most consistent overall performance by combining continuous-time inductive bias with input-conditioned gating mechanisms that enhance stability and predictive accuracy. In contrast, the CT-LTC variant, despite its theoretical appeal of calendar-aware temporal dynamics, does not yield significant gains; its uniform application of calendar-derived step sizes appears to oversmooth the discontinuous, jump-driven returns characteristic of natural gas markets. The standard LSTM achieves competitive directional accuracy but lacks meaningful explanatory power, as indicated by near-zero or negative $R^{2}$ values. This distinction underscores a key finding: architectures that predict return direction effectively do not necessarily capture the magnitude and structure of return dynamics.

Overall, this study demonstrates that liquid neural architectures, particularly Hybrid CfC and LTC, provide a reliable framework for short-term forecasting in volatile commodity markets. Improved predictive accuracy in this setting can support operational decisions for utilities, pipeline operators, and energy producers, inform trading strategies, and strengthen risk management during periods of elevated volatility. More broadly, the results confirm that integrating adaptive temporal mechanisms into neural architectures offers a principled approach to modeling complex financial time series, especially when the assumptions of smooth evolution and stationarity are violated.

\subsection{Future Work}
The most consequential limitation of the current framework is the absence of physical market fundamentals. The feature set relies on financial and equity market proxies, whereas natural gas prices are ultimately determined by physical supply and demand dynamics. EIA weekly storage reports, pipeline flows, LNG export volumes, and rig activity are among the main price forming drivers in the US natural gas market, and their exclusion limits the models' ability to anticipate structurally driven price movements. Incorporating these variables alongside temperature driven consumption data, which captures the strong seasonal and regime dependent demand patterns that financial proxies cannot represent, is the highest priority extension for improving signal quality.

Beyond the feature set, the evaluation framework has two notable gaps.
First, the restriction to one-step-ahead forecasts, while appropriate for
controlled architectural comparison, does not address the multi-day
horizons most relevant to practical applications such as hedging and
procurement. Whether the advantage of Hybrid CfC and LTC persists under
multi-step forecasting, or erodes as forecast uncertainty accumulates,
remains an open question. Second, the absence of classical time-series
benchmarks such as ARIMA and GARCH makes it difficult to attribute
performance gains specifically to nonlinear learning or continuous-time
dynamics, rather than to the adaptive re-estimation that any
rolling-window method provides.

Taken together, these gaps suggest that enriching the information set and
broadening the evaluation framework offer greater marginal returns than
further increasing model complexity. The strong performance of Hybrid CfC
and LTC indicates that adaptive temporal dynamics can already extract
meaningful structure from the available data; the priority for future work
is therefore to supply those dynamics with richer, more price-relevant
inputs and to validate their advantages across a wider range of forecasting
conditions.

\renewcommand{\refname}{Reference}

\bibliographystyle{IEEEtran}
\bibliography{references}
\end{document}